\documentclass[conference]{IEEEtran}
\IEEEoverridecommandlockouts
\usepackage[hyphens]{url}
\usepackage[hidelinks]{hyperref}
\hypersetup{breaklinks=true}
\usepackage{cite}
\usepackage{amsmath,amssymb,amsfonts}
\usepackage{graphicx}
\usepackage{subcaption}
\usepackage{textcomp}
\usepackage{xcolor}
\usepackage{url}
\def\BibTeX{{\rm B\kern-.05em{\sc i\kern-.025em b}\kern-.08em
    T\kern-.1667em\lower.7ex\hbox{E}\kern-.125emX}}

\usepackage{algorithm}
\usepackage{algorithmicx}
\usepackage[noend]{algpseudocode}% http://ctan.org/pkg/algorithmicx
\usepackage[absolute]{textpos}
\DeclareMathOperator*{\argmax}{argmax}
\DeclareMathOperator*{\argmin}{argmin}

\begin{document}

\begin{textblock}{12}(2,0.3)
	\noindent Please cite as follows: K. Malialis, D. Papatheodoulou, S. Filippou, C. G. Panayiotou, M. M. Polycarpou. Data Augmentation On-the-fly and Active Learning in Data Stream Classification. In IEEE Symposium Series on Computational Intelligence (SSCI), 2022.
\end{textblock}

\bstctlcite{IEEEexample:BSTcontrol}
\title{Data Augmentation On-the-fly and Active Learning in Data Stream Classification
\thanks{This work has been supported by the European Research Council (ERC) under grant agreement No 951424 (Water-Futures), by the European Union’s Horizon 2020 research and innovation programme under grant agreements No 883484 (PathoCERT) and No 739551 (TEAMING KIOS CoE), and from the Republic of Cyprus through the Deputy Ministry of Research, Innovation and Digital Policy.}
}

\author{\IEEEauthorblockN{Anonymous Authors}}

\author{
	\IEEEauthorblockN{
		Kleanthis Malialis\textsuperscript{a, b},
		Dimitris Papatheodoulou\textsuperscript{a},
		Stylianos Filippou\textsuperscript{a},\\
		Christos G. Panayiotou\textsuperscript{a, b} and
		Marios M. Polycarpou\textsuperscript{a, b}
	}
	\IEEEauthorblockA{
		\textsuperscript{a} \textit{KIOS Research and Innovation Center of Excellence}\\
		\textsuperscript{b} \textit{Department of Electrical and Computer Engineering}\\
		\textit{University of Cyprus}, Nicosia, Cyprus\\
		Email: \{malialis.kleanthis, papatheodoulou.dimitris, filippou.stylianos, christosp, mpolycar\}@ucy.ac.cy\\
		ORCID: \{0000-0003-3432-7434, 0000-0002-1922-1781, 0000-0001-8123-8309, 0000-0002-6476-9025, 0000-0001-6495-9171\}
	}
}

\maketitle

\begin{abstract}
There is an emerging need for predictive models to be trained on-the-fly, since in numerous machine learning applications data are arriving in an online fashion. A critical challenge encountered is that of limited availability of ground truth information (e.g., labels in classification tasks) as new data are observed one-by-one online, while another significant challenge is that of class imbalance. This work introduces the novel Augmented Queues method, which addresses the dual-problem by combining in a synergistic manner online active learning, data augmentation, and a multi-queue memory to maintain separate and balanced queues for each class. We perform an extensive experimental study using image and time-series augmentations, in which we examine the roles of the active learning budget, memory size, imbalance level, and neural network type. We demonstrate two major advantages of Augmented Queues. First, it does not reserve additional memory space as the generation of synthetic data occurs only at training times. Second, learning models have access to more labelled data without the need to increase the active learning budget and / or the original memory size. Learning on-the-fly poses major challenges which, typically, hinder the deployment of learning models. Augmented Queues significantly improves the performance in terms of learning quality and speed. Our code is made publicly available.
\end{abstract}


\section{Introduction}
Nowadays, in numerous applications information is becoming available in an online or streaming fashion. Applications include monitoring of critical infrastructure systems (e.g., leakage detection in water distribution networks \cite{kyriakides2014intelligent}), security (e.g., spam filtering \cite{wang2018systematic}), environmental monitoring \cite{ditzler2015learning}, and recommender systems \cite{ditzler2015learning}. Deploying online learning algorithms in real-world applications to train predicting models on-the-fly is impeded by a series of open challenges and problems.

A key challenge is the label availability for a classification task as data are arriving online. Acquiring the labels can be expensive or even impossible in some real-time tasks. Class imbalance is another challenge, which refers to the problem of having a skewed data distribution \cite{wang2018systematic}. Imbalance may render a traditional learning model ineffective as its predictive power on minority class examples declines significantly.

To cope with learning from limited-labelled data, we focus on active learning, a paradigm in which the model queries an oracle (typically, a human expert) for the ground truth information of selected examples \cite{settles2009active}. Active learning is part of several successful industrial systems, such as, Google’s method for labelling malicious advertisements \cite{sculley2011detecting}, and NVIDIA’s \cite{nvidia} and Tesla’s \cite{tesla} methods for their autonomous vehicles.

A recently proposed method called ActiQ uses online active learning in synergy with a multi-queue memory \cite{malialis2020data}. It was shown to address the aforementioned challenges, however, it typically performs well when the queue lengths are sufficiently large. Overall, the ActiQ method has the following limitations, which we try to address in this paper: (i) In online environments where data are sampled from a long, potentially infinite, sequence, it is impractical to assume that previous examples whose label had been requested, will always be available; (ii) To initialise the multi-queue system, it requires a large amount of historical labelled data; (iii) Its memory requirements are high. As a result, serious deployment concerns are raised.
%(ii) Under nonstationary conditions, large queues will contain more ``obsolete'' examples. To this end, incremental learning algorithms would be heavily affected;
This work makes the following key contributions.
\begin{enumerate}
    \item We propose the Augmented Queues method which significantly extends ActiQ by incorporating on-the-fly augmentation in synergy with active learning. It overcomes ActiQ's limitations (listed above), and we show its applicability using image and time-series augmentations.
    
    \item We perform an extensive study in which we include two active learning methods, image and time-series datasets, two types of neural networks (standard, VGG). We also examine the roles of the active learning budget, memory size, and imbalance level. Models using Augmented Queues have access to more labelled data without the need to increase the budget and / or the original memory size. Augmented Queues significantly improves the learning quality and speed. Our code is made available\footnote{\url{https://github.com/kmalialis/augmented_queues}}.
\end{enumerate}

The paper is organised as follows.
Section~\ref{sec:background} provides the background material. Related work is presented in Section~\ref{sec:related_work}. Augmented Queues is described in Section~\ref{sec:proposed}. The experimental setup and results are presented in Sections~\ref{sec:experimental_setup} and \ref{sec:experimental_resuts} respectively. We conclude in Section~\ref{sec:conclusion}.

\section{Preliminaries}\label{sec:background}
\textbf{Online} learning considers a data generating process that provides at each time $t$ a sequence of examples $S = \{(x^t,y^t)\}_{t=1}^T$, where the number of steps is denoted by $T \in [1, \infty)$ and data are typically sampled from a long, potentially infinite, sequence \cite{ditzler2015learning}. The examples are drawn from an unknown probability distribution $p^{t}(x,y)$, where $x^t \in \mathbb{R}^d$ is a $d$-dimensional vector in the input space $X \subset \mathbb{R}^d$, $y^t \in \{1, ..., K\}$ is the class label in the target space $Y \subset \mathbb{Z}^+$, and $K \geq 2$ is the number of classes.

An online classifier receives a new instance $x^t$ at time $t$ and makes a prediction $\hat{y}^t$ based on a concept $h: X \to Y$. In \textbf{online supervised} learning, the classifier receives the true label $y^t$, its performance is evaluated using a loss function and is then trained based on the loss incurred. The process is repeated at each step. In online applications, however, the label cannot be typically provided, as it's either impossible (e.g. in real-time applications) or expensive and thus impractical.

To address this issue, an alternative paradigm is active learning \cite{settles2009active}, which deals with strategies to selectively query for labels from a human expert according to a pre-defined ``budget'' $B \in [0,1]$, for example, $B=0.3$ means that $30\%$ of the arriving instances can be labelled. A budget spending mechanism must ensure that the labelling spending $b \in [0,1]$ does not exceed the allocated budget.

In \textbf{online active} learning \cite{zliobaite2013active}, a classifier is built that receives a new instance $x^t$ at time $t$. At each time step the classifier calculates the prediction probability $\hat{p}(y | x^t)$. The classifier outputs the best prediction probability $h(x^t) = \max_y \hat{p}(y|x^t)$ and the predicted class $\hat{y}^t = \argmax_y \hat{p}(y|x^t)$. A given active learning strategy $\alpha : X \to \{False,True\}$ decides if the true label $y^t$ is required, which is assumed that the oracle will provide. The classifier is evaluated using a loss function and is then trained based on the loss incurred. Typically, training occurs only when $\alpha(x^t) = True$.

% A key challenge encountered in some online applications is that of data \textbf{nonstationarity} \cite{ditzler2015learning, gama2014survey, lu2018learning}, typically caused by \textbf{concept drift}, which represents a change in the joint probability. The drift between steps $t_i$ and $t_j$, where $i \ne j$, is defined as follows:
% \begin{equation}
% \quad p^{t_i}(x,y) \neq p^{t_j}(x,y)
% \end{equation}

One major challenge encountered in some streaming applications is the presence of infrequent events, also known as \textbf{class imbalance} \cite{he2008learning, wang2018systematic}. It occurs when at least one class is under-represented, thus constituting a minority class. In binary classification, imbalance is defined as follows:
\begin{equation}
\exists y_0, y_1 \in Y \quad p^t(y=y_0) \gg p^t(y=y_1),
\end{equation}
\noindent where $y_1$ represents the minority class.

\section{Related Work}\label{sec:related_work}
Online supervised learning methods that address imbalance are grouped as \cite{wang2018systematic} (i) resampling methods, e.g., Oversampling-based Online Bagging (OOB) \cite{wang2015resampling}, Adaptive REBAlancing (AREBA) \cite{malialis2020online, malialis2018queue}, and Hybrid-AREBA (HAREBA) \cite{malialis2022hybrid}; and (ii) cost-sensitive learning methods, e.g., CSOGD \cite{wang2014cost}. Despite their effectiveness, they rely on continual supervision. This work focuses on online active learning, and augmentation.

\subsection{Online active learning}

\subsubsection{Querying strategies}
The most widely used active learning strategy is uncertainty sampling, where the learner queries the most uncertain instances, which are typically found near the decision boundary \cite{lewis1994sequential}. Most existing strategies assume that the training set $U \subset X$ is already available (offline active learning) \cite{cohn1994improving}. One way to measure uncertainty \cite{settles2009active} is to first find the instance $x_q$ with the least confident best prediction:
\begin{equation}
x_q = \argmin_{x \in U} h(x)
\end{equation}
\noindent where $h(x) = \max_y \hat{p}(y|x)$ and request its label if it satisfies the following condition:
\begin{equation}
h(x_q) < \theta,
\end{equation}
where $\theta$ is a threshold which is typically fixed. Work on online active learning is limited. The arriving $x^t$ is queried if:
\begin{equation}
h(x^t) < \theta,
\end{equation}
\noindent where $h(x^t) = \max_y \hat{p}(y|x^t)$ and $\theta$ is a fixed threshold. This is called a \textit{fixed uncertainty sampling} strategy \cite{zliobaite2013active}.

This strategy may not perform well if the threshold is set incorrectly, or if the classifier learns enough so that the uncertainty remains above the fixed threshold most of the time. In \cite{zliobaite2013active} a variable uncertainty sampling strategy is proposed, which uses randomisation to ensure that the probability of labelling any instance remains above zero. This is termed \textit{randomised variable uncertainty sampling (RVUS)} strategy and the threshold is modified as follows:
\begin{equation}\label{eq:strategy}
\theta =
\begin{cases}
\theta (1 - s) & \text{if } h(x^t) < \theta_{rdm} \text{ \# request label}\\
\theta (1 + s) & \text{if } h(x^t) \geq \theta_{rdm} \text{ \# don't request}\\
\end{cases}
\end{equation}
\noindent where $s$ is a step size parameter, $\theta_{rdm} = \theta * \eta$ where $\eta$ follows a Normal distribution $\eta \sim N(1,\delta)$ with a standard deviation of $\delta$.

A recently proposed method called ActiQ \cite{malialis2020data}, has achieved state-of-the-art results in imbalanced scenarios by combining the RVUS strategy with a multi-queue data storage. In this work we propose the Augmented Queues method, which significantly extends ActiQ to allow data augmentation.

For a comprehensive review the interested reader is directed towards \cite{settles2009active}. Lastly, we adopt the widely-used budget spending mechanism from \cite{zliobaite2013active}. Due to space constraints we direct the interested reader to \cite{zliobaite2013active}, and to our released code.

% \subsubsection{Budget spending mechanisms}\label{sec:budget_spending}
% By counting the \textit{exact} labelling spending \cite{zliobaite2013active}, the contribution of every next label diminishes over time. To address this issue we could count the \textit{exact} labelling spending over a sliding window $b^t= \frac{u^t_w}{w}$, where $u^t_w$ is the number of instances queried within the window $w$. However, this contradicts the incremental learning concept as it needs to store previously labelling decisions. Alternatively, we could approximate the labelling spending $\hat{b}^t= \frac{\hat{u}^t_w}{w}$ by approximating the number of instances queried in the window as follows: $\hat{u}^t_w = \lambda \hat{u}^{t-1}_w + a(x^t)$, where $\lambda = \frac{w-1}{w}$ and $a(x^t)$ is a Boolean value that indicates if the true label for $x^t$ is queried or not. It was proved that $\hat{b}$ is an unbiased estimate of $b$ \cite{zliobaite2013active}. We adopt this approach in our work.

\subsection{Data augmentation}\label{sec:data_augmentation}
Augmentation is applied to a dataset to expand its size by artificially creating variations of the data \cite{shorten2019survey}. It enhances the diversity of the dataset which could improve the learning performance, and improve generalisation. %It can also combat overfitting as with the addition of augmented examples, algorithms typically tend to generalise to new scenarios of unseen data.

% \begin{figure}[t!]
% 	\centering
% 	\includegraphics[scale=0.65]{augms.png}
% 	\caption{Examples of augmentation transformations on the original image on the left. Top row: brightness change, height shift, random erase. Bottom row: rotation, width shift, zoom.}
% 	\label{fig:augms}
% \end{figure}

\subsubsection{Image augmentation} Such techniques include:
%There are various augmentation techniques that can be applied to image problems, however, some of the techniques cannot be used for every dataset. For example, on the MNIST dataset, a collection of hand-written digit images, variations such as horizontal or vertical flip would be misleading in cases where the resulting augmented image is similar to an image of a different class, such as 2 and 5 or 6 and 9.
%\par There are several types of image augmentations techniques including geometric transformations, colour-space augmentations, kernel filters, mixing images, random erasing, feature space augmentation, adversarial training, generative adversarial networks, neural style transfer, and meta-learning. The most frequently used techniques are positional and colour transformations due to their ease of use.

\textbf{Geometric techniques} \cite{taylor2018improving} alter the pixel positions of images; some examples include scaling, cropping, flipping, padding, rotation, and image translation.

\textbf{Colour-space augmentations} \cite{shorten2019survey} alter the colour properties of images by changing their pixel values, such as, changes to brightness, contrast, saturation, and hue of the images.

\textbf{Random erasing} \cite{zhong2020random}
%was initially inspired by the concept of dropout regularisation, with the difference that instead of being embedded into the network architecture it alters the input data space.
helps to prevent overfitting as it forces the model to learn more descriptive features by cutting-off random patches from the images. Patches can be masked with pixel values of 0s, 255s or mean pixel values, nonetheless, it was shown that random noise masking yields the best results.

%Fig.~\ref{fig:augms} shows examples of image transformations. For more image augmentation techniques we direct the interested reader towards this excellent survey article \cite{shorten2019survey}.

%Other data augmentation techniques include image mixing \cite{inoue2018data}, kernel filters \cite{kang2017patchshuffle}, adversarial training \cite{goodfellow2014explaining}, generative modelling \cite{bowles2018gan}, and style transfer \cite{jackson2019style}.

%The \textbf{image mixing} \cite{inoue2018data} transformation is considered to be counter-intuitive compared to the previously mentioned ones, as it averages the pixel values of a pair of images which does not look like a useful technique to a human observer, however, it has demonstrated great results and it proved to be an effective augmentation strategy.
% \textbf{Kernel filters} \textcolor{red}{[REF]} are also a popular technique that can be applied to sharpen or blur images.
% More advanced techniques include adversarial training, generative modelling and style transfer. These techniques rely on a network that is trained on the image dataset and learns to generate synthetic instances that retain similar characteristics to the original images.

\subsubsection{Time-series augmentation} Such techniques include:

\textbf{Window Slicing} is a time domain transformation method, for extracting slices from a time series, and assigning the same class $y$ as the time series used for slice extraction \cite{guennec2016data}.

\textbf{Time Warping} \cite{umtt2017data} is a time domain transformation method, which aims to disrupt a pattern in the temporal domain either by using a randomly located fixed window \cite{guennec2016data} or by using a smooth warping path.

%For more time-series augmentation techniques we direct the interested reader towards this excellent survey article \cite{iwana2020empirical}.

\section{Augmented Queues}\label{sec:proposed}
\begin{figure*}[t!]
	\centering
	\includegraphics[scale=0.45]{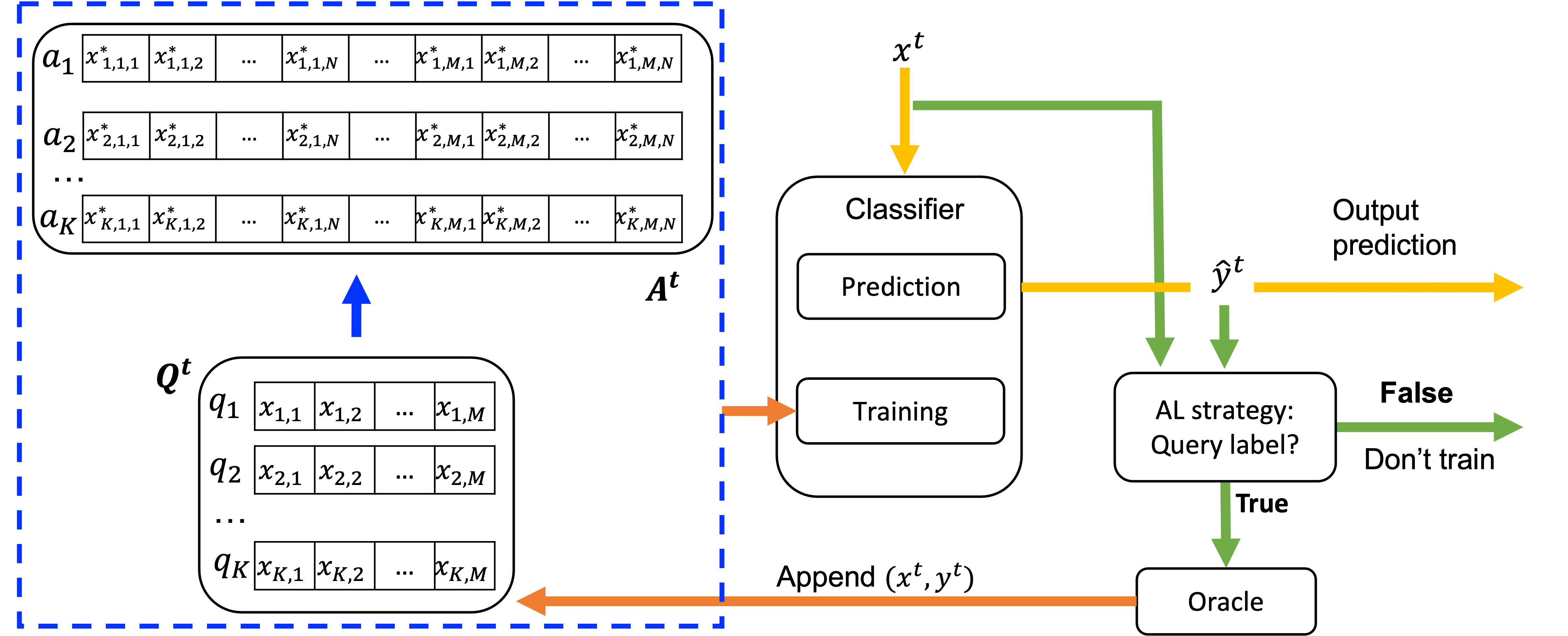}
	\caption{The overview of the proposed method, Augmented Queues}
	\label{fig:overview}
\end{figure*}

The overview of the Augmented Queues method is shown in Fig~\ref{fig:overview}. At any time $t$, the classifier observes an arriving instance $x^t$, and then provides a prediction $\hat{y}^t$; this is shown in yellow colour. If the active learning strategy does not request the ground truth, i.e., the class label, no training is performed and the algorithm waits for the next arriving example. If the strategy requests the ground truth, this is provided by an oracle (typically, a human expert) as shown in green colour. The example ($x^t,y^t$) is then appended to the relevant queue in $Q^t$. Before training, the data augmentation process is initiated to create the augmented queues $A^t$. The neural network is then trained using both $Q^t$ and $A^t$. This is shown in orange colour. We describe below each individual element, as well as a discussion on the computational aspects of the method.

\subsection{Individual elements}

\textbf{Multi-queue memory:}
The method uses multiple first-in-first-out (FIFO) queues which will be populated by instances queried by the active learning strategy. At any time $t$ we maintain a set of $K$ queues, one for each class as follows:
\begin{equation}
Q^t = \{q^t_1, q^t_2,..., q^t_K\} = \{q^t_c\}^K_{c=1},
\end{equation}
\noindent where $K \geq 2$ is the number of classes. All queues are of the same capacity $M$ and a queue corresponding to class $c$ is defined as follows:
\begin{equation}
q^t_c = \{x_{c,1}, x_{c,2}, ..., x_{c,M}\} = \{x_{c,i}\}^M_{i=1},
\end{equation}
\noindent where $x_{c,i} \in \mathbb{R}^d$, and for any two $x_{c,i}, x_{c,j} \in q^t_c$ such that $j > i$, $x_{c,j}$ has been observed more recently in time.

As in the original \textit{ActiQ} work, we assume the initial availability of $M$ labelled examples per class. Balanced queues ensure robustness to class imbalance; the assumption is not needed for problems in which imbalance does not exist, as the queues will be populated at a similar rate. As this may be difficult to have in practise, the purpose of this work is to restrict $M$ to a very small number e.g. up to ten, and then apply data augmentation. We argue that for the vast majority of applications this is realistic and practical. 

\textbf{Augmented memory:} Let $F = \{f_o\}_{o=1}^{|F|}$ be a set of $|F|$ available transformation functions $f_o: X \rightarrow X$, such that, an original example $x$ is augmented to $x^* = f_o(x)$. The number of transformations $|F|$ is task-dependent, for instance, like the ones described in Section~\ref{sec:data_augmentation}. Augmentation is initiated every time before the model is trained, that is, when the active learning strategy receives a class label. Let's define the augmented multi-queue memory as follows:
\begin{equation}\label{eq:augmented_queues}
A^t = \{a^t_1, a^t_2,..., a^t_K\} = \{a^t_c\}^K_{c=1},
\end{equation}
\noindent where $K \geq 2$ is the number of classes.

Each augmented queue $a^t_c \in \mathbb{R}^{M \times N}$ has capacity $|a^t_c| = M \times N$ where $c$ is the class and $N$ is the number of augmentation transformations per example. It is defined as:
\begin{equation}
a^t_c = \{ x^*_{c,i} \in \mathbb{R}^N | \forall i \in [1,M] \}
\end{equation}
\noindent where $x^*_{c,i}$ contains all $N$ augmented examples generated from the original example $x_{c,i}$ as follows:
\begin{equation}
x^*_{c, i} = \{ x^*_{c, i, j} = f_o(x_{c,i}) \}_{j=1}^N
\end{equation}
\noindent where each time the transformation function is selected randomly from the set of functions $f_o \in_R F$.

\textbf{Class prediction}: The neural network predicts the class of each arriving instance $x^t$ as shown below. The multi-queue memory is not used in the prediction process.
\begin{equation}\label{eq:actiq_predict}
\hat{y}^t = \argmax_{y \in \{1, ..., K\}} \hat{p}(y | x^t)
\end{equation}

\textbf{Active learning strategy}: The proposed Augmented Queues method, like ActiQ, uses the RVUS \cite{zliobaite2013active} active learning strategy as shown in~Eq. (\ref{eq:strategy}).

\textbf{Incremental learning}: Recall that the neural network is trained only when the active learning strategy receives the class label. Prior the training at time $t$, the data augmentation process is initiated to create the augmented queues $A^t$ from the original queues $Q^t$. We then merge the two queues as follows: ${Q^*}^t = Q^t \cup A^t$. The output of the neural network is provided to a softmax output unit and its cost function is: 
\begin{equation}\label{eq:actiq_cost}
	J^t = \frac{1}{|{Q^*}^t|} \sum_{x_i \in {Q^*}^t} l(y_i, \hat{P}_{x_i} ),
\end{equation}
\noindent where $y_i \in \{1, ..., K\}$  is the ground truth and $\hat{P}_{x_i} =  \{\hat{p}(1 | x_i), ..., \hat{p}(K | x_i)\}$ are the prediction probabilities for each class. The loss function used is the cross entropy $l(y, \hat{P}_x) = - \sum^{K}_{c=1} \mathbb{I}_{y = c} \: \log \: \hat{p}(c | x)$, where $\mathbb{I}_{condition}$ is the identity function that returns $1$ if the condition is satisfied. The neural network $h$ will be updated incrementally based on the cost incurred, that is, $h^t = h^{t-1}.train(J^t)$. The pseudocode of the Augmented Queues method is provided in Algorithm~\ref{alg:method}.

\begin{algorithm}[]
	\caption{Augmented Queues}
	\label{alg:method}
	\begin{algorithmic}[1]
		
		\Statex \textbf{Input:}
		\State $a$: active learning strategy
		\State $B$: labelling budget
		\State $K$: number of classes
		\State $M$: memory / queue size
		\State $D$: initial labelled examples \Comment $|D| = K \times M$
		
		\Statex \textbf{Initialisation:}
		\State init queues $Q^0 = FIFOs(capacity=M, init=D)$
		\State create model $h^0$
		\State init budget expenses $b^0 = 0$
		
		\Statex \textbf{Main:}
		\For{each time step $t \in [1, \infty)$}
		\State receive instance $x^t \in \mathbb{R}^d$
		\State predict class $\hat{y}^t$ using Eq.~(\ref{eq:actiq_predict})
		
		\State $Q^t = Q^{t-1}$
		\State $h^t = h^{t-1}$
		\If{$b^{t-1} < B$}\Comment expenses within budget
		\If{$a(x^t, h(x^t)) == True$}\Comment AL Eq. (\ref{eq:strategy})
		\State receive true label $y^t$
		\State append example $Q^t = Q^{t-1}.append(x^t,y^t)$
		\State create augmented queues $A^t$ using Eq. (\ref{eq:augmented_queues})
		\State calculate cost $J^t$ using Eq.~(\ref{eq:actiq_cost})
		\State incremental training $h^t = h^{t-1}.train(J^t)$
		\EndIf
		\EndIf
		
		\State update budget expenses $b^t$
		\EndFor
		
	\end{algorithmic}
\end{algorithm}

Augmented Queues is robust to imbalance due to the \textit{separate} and \textit{balanced} queues per class. Propagating previously observed examples in the most recent training set is a form of oversampling. The augmentation is performed in such as way that $A^t$ also contains separate and balanced queues per class.

\subsection{Computational aspects}
In online or streaming environments where data are arriving from a long, potentially infinite, sequence of data, it is unrealistic to expect that all previously observed examples will always be available during learning. Therefore, a learning model should use no more than a fixed amount of memory for any storage \cite{gama2014survey}. The multi-queue memory $Q^t$ has a fixed size of $Q^t = K \times M$. As discussed, the queue length is kept to a minimum (e.g., $M=10$) while the number of classes K is task-dependent. The augmented memory $A^t$ is fixed as well with size $|A^t| = N \times |Q^t|$, where $N$ is the number of augmentation transformations per example. Importantly, it reserves memory only at training times.

Predicting the class of an arriving instance $x^t$ only requires a forward propagation pass of the neural network. Recall that the multi-queue memory is not used in the prediction process.

Training is only performed at certain times according to the budget, that is, the model is updated every time a class label has been received. Following the recommendation by \cite{malialis2020data} to avoid overfitting, the model is updated once at each training time, i.e., the number of epochs is set to 1. This is a parameter of the optimiser (e.g., gradient descent) that corresponds to a one-pass over the entire batch (or each mini-batch) of the data.

\section{Experimental Setup}\label{sec:experimental_setup}

\subsection{Datasets}
\textbf{MNIST} \cite{lecun1998gradient}: This widely used dataset is a collection of images of handwritten digits. Each image has a 1-colour channel (monochrome), and it depicts a digit between ``0'' to ``9'' (10 classes), which is centred in a $28\times28$ pixel-sized box. MNIST, typically, serves as a benchmark dataset for image classification, however, it is important to note that it can stress test online / streaming learning algorithms due to its high-dimensionality of 784 features. To examine the effect of imbalance, we create three variations of MNIST and focus on the challenging case of multi-minority scenarios \cite{wang2016dealing}:
\begin{itemize}
    \item MNIST-balanced: It refers to balanced scenarios with 5000 arriving examples per class.

    \item MNIST-imbalanced10: It refers to scenarios with 10\% imbalance. Digit ``0'' is the majority class from which 5000 examples arrive. The rest are minority classes, with 500 arriving examples per class.

    \item MNIST-imbalanced1: It refers to scenarios with 1\% imbalance, with 50 arriving examples per minority class.
\end{itemize}

\textbf{Two Patterns} \cite{Geurts2002ContributionsTD}: A simulated time series dataset in which each class represents the presence of two patterns in a definite order, that describes upward and downward steps defined by time functions presented in \cite{Geurts2002ContributionsTD}. There are 4 classes and 5000 samples as follows: down-down class with 1306 cases, up-down with 1248 cases, down-up with 1245 cases and up-up with 1201 cases. The number of features is 128.

\textbf{uWave Gesture Library Z} \cite{liu2009uWave}: A time series dataset for a set of eight simple gestures generated from accelerometers using the Wii remote. The data consists of the Z coordinates of each motion. There are 8 classes with total 3582 samples, each with 315 features. %A more detailed description regarding the collection of the data can be found in \cite{liu2009uWave}.

%Careful consideration of the augmentation methods has been made, as some of them cannot be applied effectively to every dataset. On MNIST, for example, variations generated by horizontal or vertical flips would be misleading in cases where the resulting augmented image looks more similar to an original image of a different class. As a result, we have avoided using some geometric transformation techniques. We have also disregarded most colour-space augmentation techniques as MNIST is in grayscale format. Therefore, we have applied scaling, rotation, image translation, changes in brightness, and random erasing (RE) patches with random noise. These transformations are applied to each candidate sample randomly, either individually or stacked (i.e., combined), to increase the variety of augmented data.

Each augmentation method' values are found in our code.

\subsection{Methods}
The active learning methods used are:

\textbf{RVUS} \cite{zliobaite2013active}: The seminal work which introduced the randomised variable uncertainty sampling strategy shown in Eq.~(\ref{eq:strategy}). It uses a neural network (described below), and it is a one-pass learner as it does not use any memory. When used with data augmentation, it is applied to the most recent example for which the oracle provided its label.

\textbf{ActiQ} \cite{malialis2020data}: A state-of-the-art method which uses the RVUS strategy and a multi-queue memory. In its original form, no data augmentation is used.

\textbf{Augmented Queues}: The proposed method shown in Fig.~\ref{fig:overview}. We will refer to it as ActiQ with data augmentation, which is applied to the memory elements as described in Section~\ref{sec:proposed}.

The neural networks used with the AL methods are:

\textbf{Standard Neural Network (NN)} \cite{bishop2006}: It is a standard fully-connected feed-forward neural network, trained using back-propagation. The hyper-parameters of NN for the MNIST, Two Patterns and uWave Gesture Library Z datasets are in our publicly available code.

\textbf{VGG} \cite{simonyan2014very}:
It is a convolutional network, distinguished by its simplicity, as it consists of a series of VGG-blocks which are a sequence of convolutional layers with padding, a non-linear activation function, and a pooling component. We focus on VGG-16 which is often used in practical applications due to its effectiveness and simplicity. It is composed of 5 blocks with a total of 13 convolutional layers and 3 fully-connected layers; its total number of parameters exceeds 100 million. The hyper-parameters of VGG are in our publicly available code. Note that the VGG will start learning from scratch, i.e., no pre-training on the ImageNet dataset will be performed.

\subsection{Performance metrics and Evaluation method}\label{sec:performance_metrics}
A popular metric which is insensitive to class imbalance \cite{he2008learning} is the geometric mean, defined as \cite{sun2006boosting}: 
\begin{equation}\label{eq:gmean}
G\text{-}mean = \displaystyle\sqrt[K]{\prod_{c=1}^K R_c},
\end{equation}
\noindent where $K$ is the number of classes, and $R_c = N_{cc} / N_c$ is the recall of class $c$ where $N_{cc}$ is the number of examples correctly classified, and $N_c$ is the total number of examples for this class. %$G\text{-}mean$ has some desirable properties as it is not only insensitive to imbalance, but it is high when all recalls are high and when their difference is small \cite{he2008learning}.
To compare learning methods in a sequential setting, we use the widely adopted \textit{prequential evaluation with fading factors} method.
%It has been proven to converge to the Bayes error when learning in stationary data \cite{gama2013evaluating}. It has the advantage of not requiring a holdout set and the learning algorithm is always tested on unseen data.
The fading factor is set to $\xi = 0.99$. In all simulation experiments we plot the prequential $G\text{-}mean$ in every time step averaged over 20 repetitions, including the error bars displaying the standard error around the mean.

\section{Experimental Results}\label{sec:experimental_resuts}
%Our study examines several factors that can affect the model's performance. These factors include the active learning budget, neural network depth, memory size, and augmentation.

\subsection{Role of the budget and the model depth}\label{sec:role_of_budget_and_model_depth}
\begin{figure}[t!]

    \begin{subfigure}[b]{0.32\columnwidth}
        \centering
        \includegraphics[width=\linewidth]{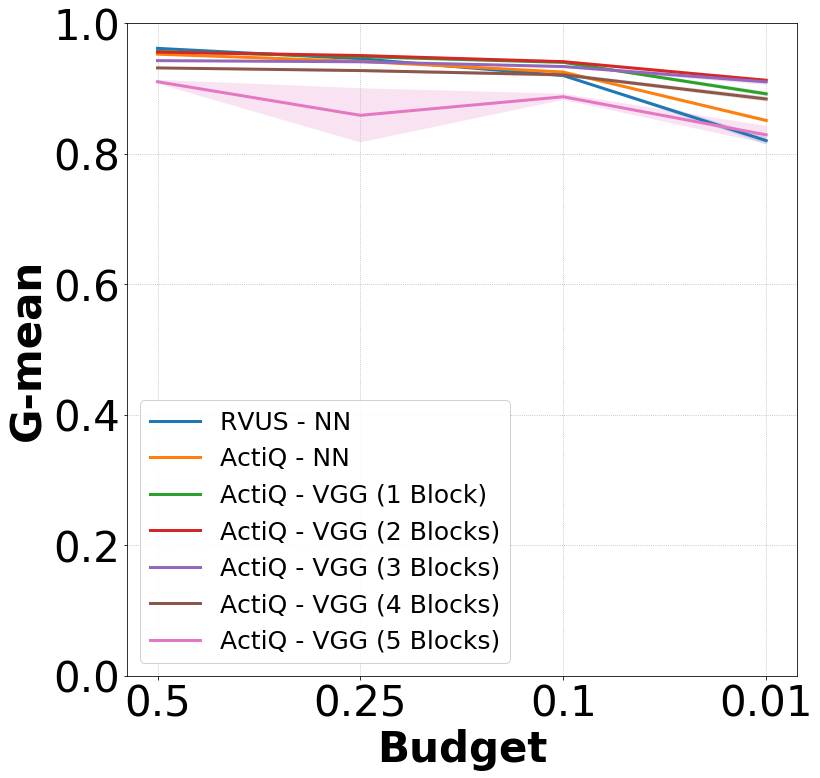}
        \caption{Class balance}
        \label{fig:MNIST_Fully_Balanced_Memory_10}
    \end{subfigure}
    \hfill
    \begin{subfigure}[b]{0.32\columnwidth}
        \centering
        \includegraphics[width=\linewidth]{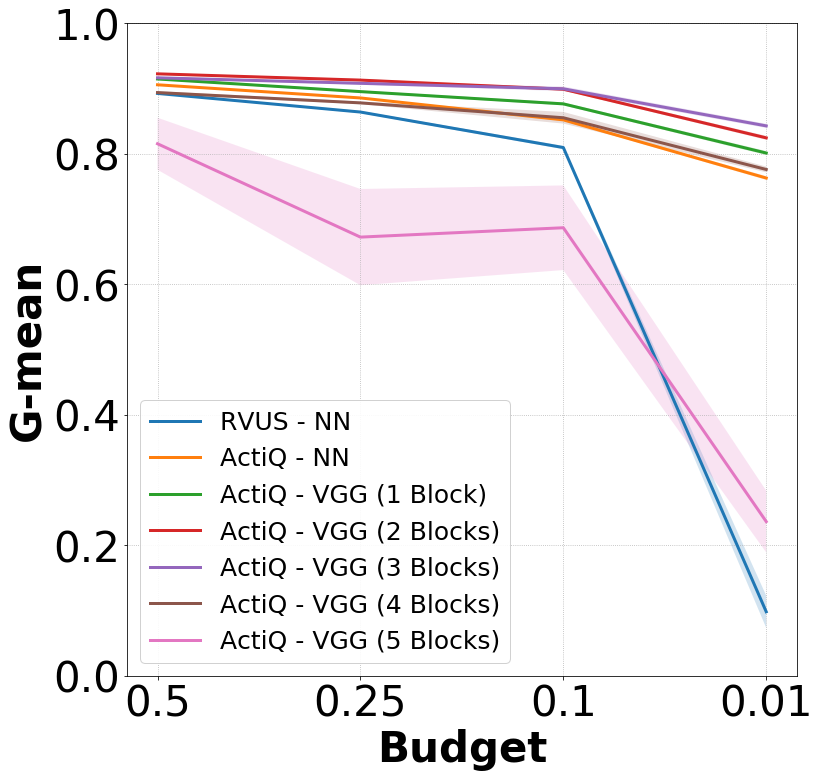}
        \caption{Imbalance 10\%}
        \label{fig:MNIST_Imbalanced_10perc_Memory_10}
    \end{subfigure}
    \hfill
    \begin{subfigure}[b]{0.32\columnwidth}
        \centering
        \includegraphics[width=\linewidth]{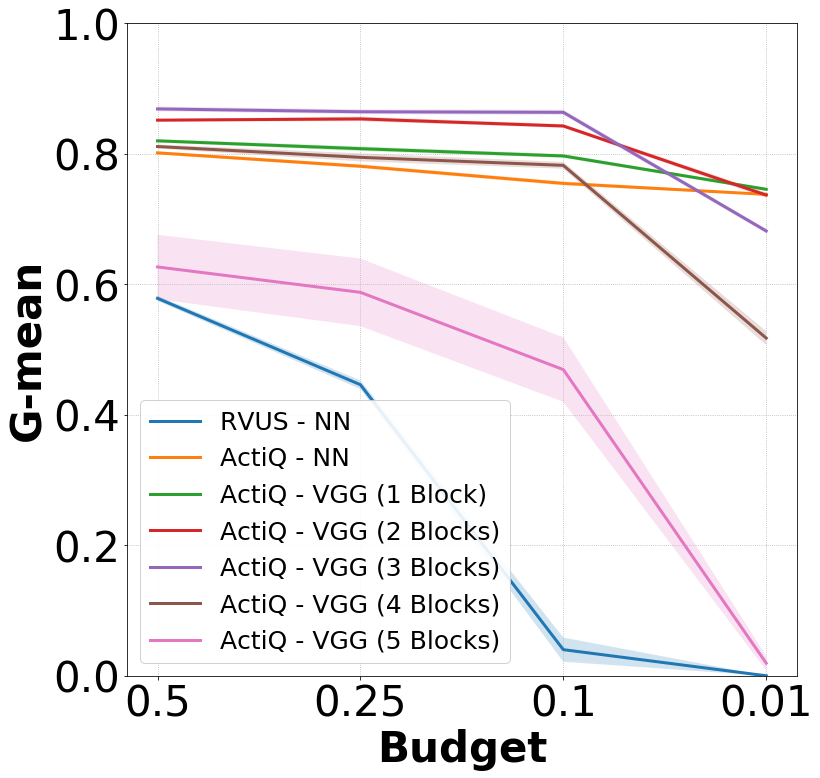}
        \caption{Imbalance 1\%}
        \label{fig:MNIST_Imbalanced_1perc_Memory_10}
    \end{subfigure}
    
\caption{The role of the budget and the model depth on the final performance ($t = 50000$, $t = 9500$, $t = 5450$) in MNIST}
\label{fig:MNIST_Fully_Balanced_Memory_1_10}
\end{figure}

These experiments examine the behaviour of NN and VGG (1-5 blocks) under various budgets. The models have a queue length of 10. Figure \ref{fig:MNIST_Fully_Balanced_Memory_10} shows the performance on MNIST-balanced, whereas Figures \ref{fig:MNIST_Imbalanced_10perc_Memory_10} and \ref{fig:MNIST_Imbalanced_1perc_Memory_10} depict the performance on MNIST with 10\% and 1\% imbalance respectively. For each dataset, we examine 4 different budgets, 0.5, 0.25, 0.1, and 0.01, and we present the results on the final performance.

In Fig. \ref{fig:MNIST_Fully_Balanced_Memory_10}, the models achieve their peak performance on budget 0.5. A similar performance is achieved on smaller budgets, e.g., 0.25, 0.1, and even 0.01. The highest performing models are the VGGs with 2 and 3 blocks, which achieve almost identical performance. The VGGs with 1 and 4 blocks perform similar to the NN, with the exception of budget 0.01 where they achieve a higher score. While RVUS-NN and ActiQ-NN have similar performances, we notice a performance drop for the former on budget 0.01. Interestingly, the worst performing models for budgets 0.5, 0.25 and 0.1 is the VGG (5 blocks), while for budget 0.01 is the RVUS NN.

In Fig. \ref{fig:MNIST_Imbalanced_10perc_Memory_10}, similarly with the balanced dataset, the models achieve the best results when the budget is 0.5. The effect of the budget is slightly more prominent, especially on budget 0.01. The VGGs with 2 and 3 blocks achieve the best performance overall, while the latter achieves the best performance on the lowest budget. Most VGG networks outperform ActiQ-NN, with the exception of VGG with 5 blocks and RVUS-NN. Regarding RVUS-NN, the model outperforms the VGG with 5 blocks on budgets 0.5, 0.25 and 0.1, but on budget 0.01 we notice a significant drop of for RVUS-NN. This establishes the RVUS-NN as the worst performing model on budget 0.01.

In Fig. \ref{fig:MNIST_Imbalanced_1perc_Memory_10}, the best performing model is the VGG with 3 blocks for most budgets, followed by VGG with 2 blocks. However, on budget 0.01, the VGG (3 blocks) performs worse than its shallower counterparts as well as the NN. As before, the deepest VGG model performs the worst along with RVUS-NN. Here, the effect of the VGG model depth is more prominent as the budget decreases. Important remarks are:
\begin{itemize}
    \item As the budget decreases, the performance drops, attributed to the fewer labelled examples, therefore, the queues are populated by a slower rate with new examples. Training also occurs more infrequently.

    \item As imbalance becomes higher, the performance drops, attributed to the fact that the minority classes are populated by a smaller rate with new examples.

    \item The role of the model type plays a key role. As expected, the convolutional network VGG outperforms the standard NN on image classification tasks, however, the model depth plays an important role. Shallow architectures have limited capacity, while deeper architectures have a larger capacity but are typically more difficult to train. For instance, VGG with 3 blocks appears to provide a good trade-off, while the original VGG (5 blocks) is even outperformed by the standard NN.
    
    \item The one-pass learner RVUS yields the worst performance due to the lack of a memory component. Interestingly, in some cases it outperformed the original VGG (5 blocks).
\end{itemize}

\begin{figure}[t!]

    \begin{subfigure}[b]{0.3\columnwidth}
        \centering
        \includegraphics[width=\linewidth]{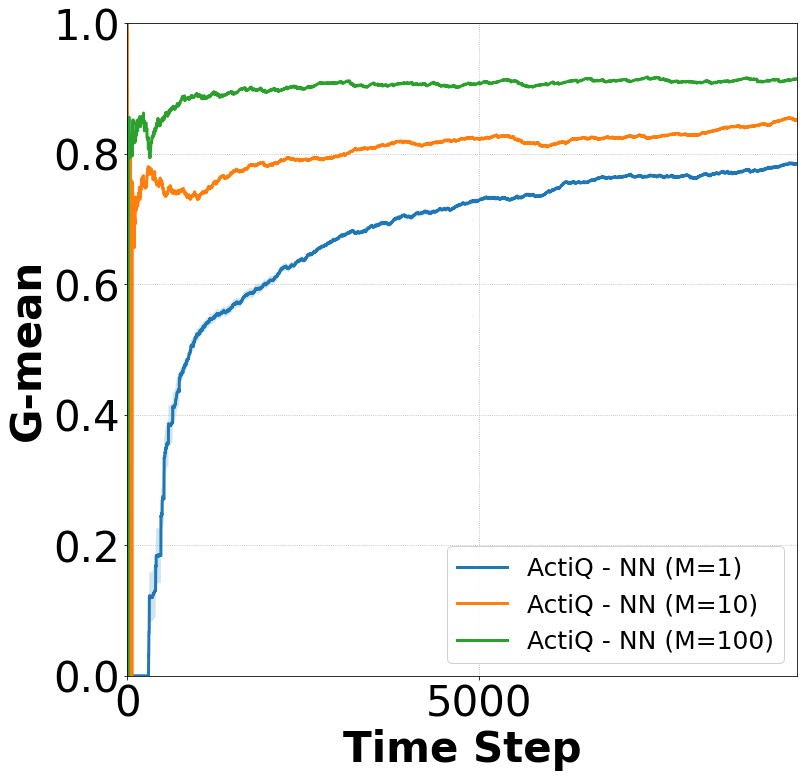}
        \caption{ActiQ-NN (10\% imbalance)}
        \label{fig:MNIST_fixed_budget_NN_imbalance_10perc}
    \end{subfigure}
    \hfill
    \begin{subfigure}[b]{0.3\columnwidth}
        \centering
        \includegraphics[width=\linewidth]{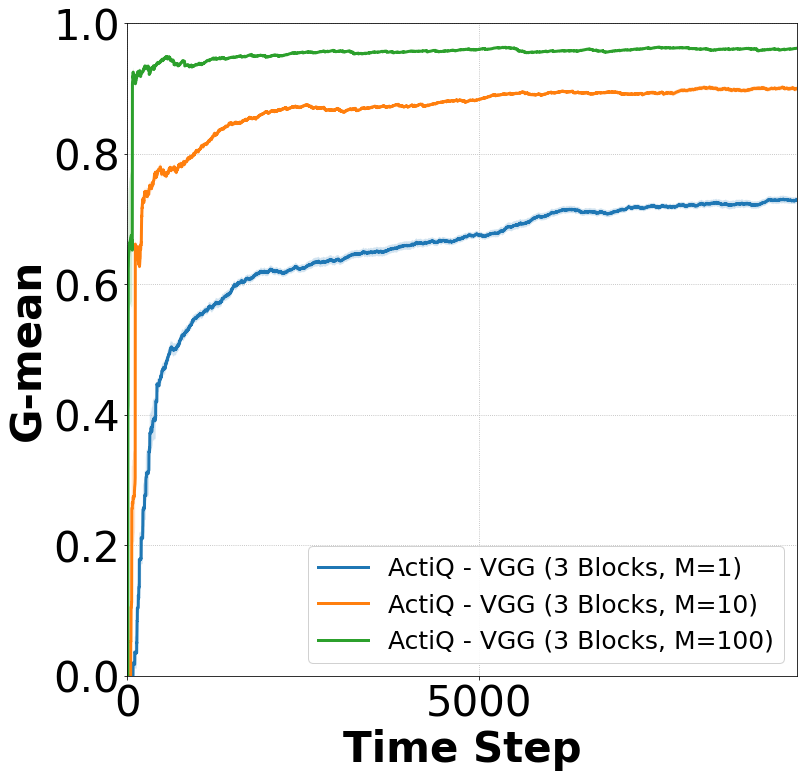}
        \caption{ActiQ-VGG (10\% imbalance)}
        \label{fig:MNIST_fixed_budget_VGG_imbalance_10perc}
    \end{subfigure}
    \hfill
    \begin{subfigure}[b]{0.3\columnwidth}
        \centering
        \includegraphics[width=\linewidth]{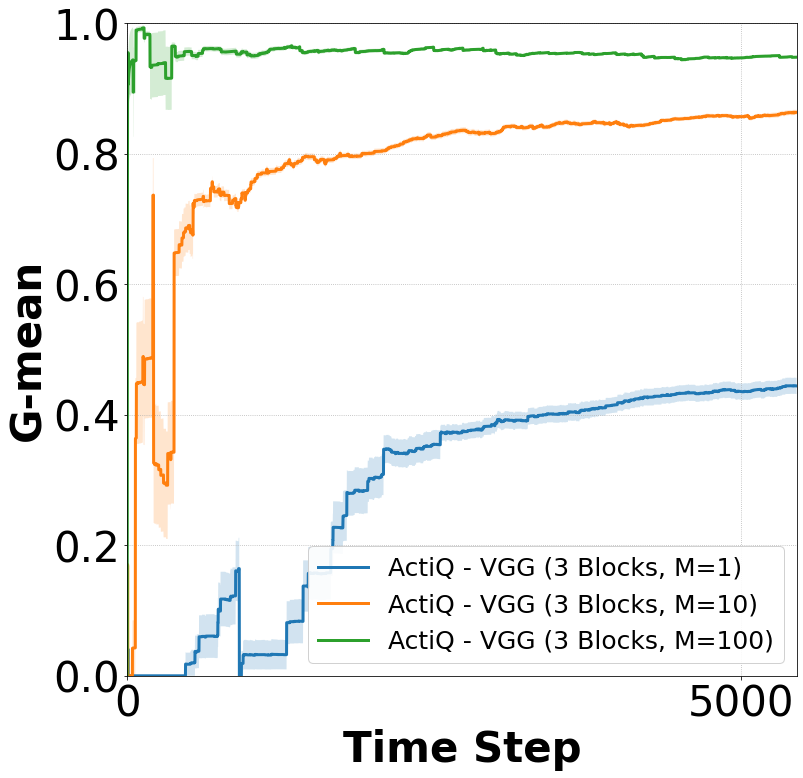}
        \caption{ActiQ-VGG (1\% imbalance)}
        \label{fig:MNIST_fixed_budget_VGG_imbalance_1perc}
    \end{subfigure}
    
\caption{The role of the memory size $M$ for ActiQ in MNIST with budget $B=10\%$.}
\label{fig:MNIST_fixed_budget_NN}
\end{figure}

\subsection{Role of the memory size}
We examine now the memory size's role; we compare NN, and VGG with 3 blocks which achieved the best overall performance in Section \ref{sec:role_of_budget_and_model_depth}. The budget is fixed to 0.1.
%which is the point where we observed the best performance from the models with the least training frequency.
In these experiments, we present the learning curves, that display the performance at different time steps. Fig.~\ref{fig:MNIST_fixed_budget_NN_imbalance_10perc} shows ActiQ-NN's performance in MNIST with 10\% imbalance. Figs.~\ref{fig:MNIST_fixed_budget_VGG_imbalance_10perc} and \ref{fig:MNIST_fixed_budget_VGG_imbalance_1perc} show ActiQ-VGG's performance in MNIST with 10\% and 1\% imbalance respectively. Important remarks are:
\begin{itemize}
 \item By increasing the memory size, the performance is significantly improved. This is attributed to the larger training set, which could particularly aid larger models.
 
 \item In severely imbalanced scenarios the role of the memory size becomes even more important. As before, this is attributed to queues corresponding to the minority classes which are populated by a smaller rate with new examples.
 
 \item We stress out, however, that in online / streaming environments, a large memory size is not desirable. This constitutes a limitation of the original ActiQ.
\end{itemize}

\subsection{Role of data augmentation}
\begin{figure}[t]

    \begin{subfigure}[b]{0.3\columnwidth}
        \centering
        \includegraphics[width=\linewidth]{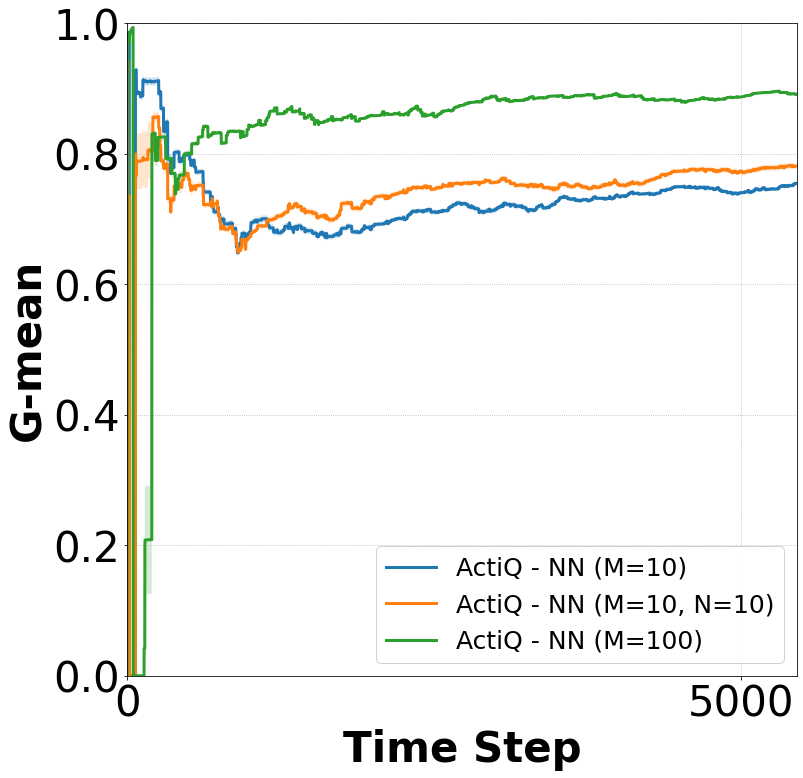}
        \caption{ActiQ-NN, $B=10\%$}
        \label{fig:nn_augs_comparison_b0.1}
    \end{subfigure}
    \hfill
    \begin{subfigure}[b]{0.3\columnwidth}
        \centering
        \includegraphics[width=\linewidth]{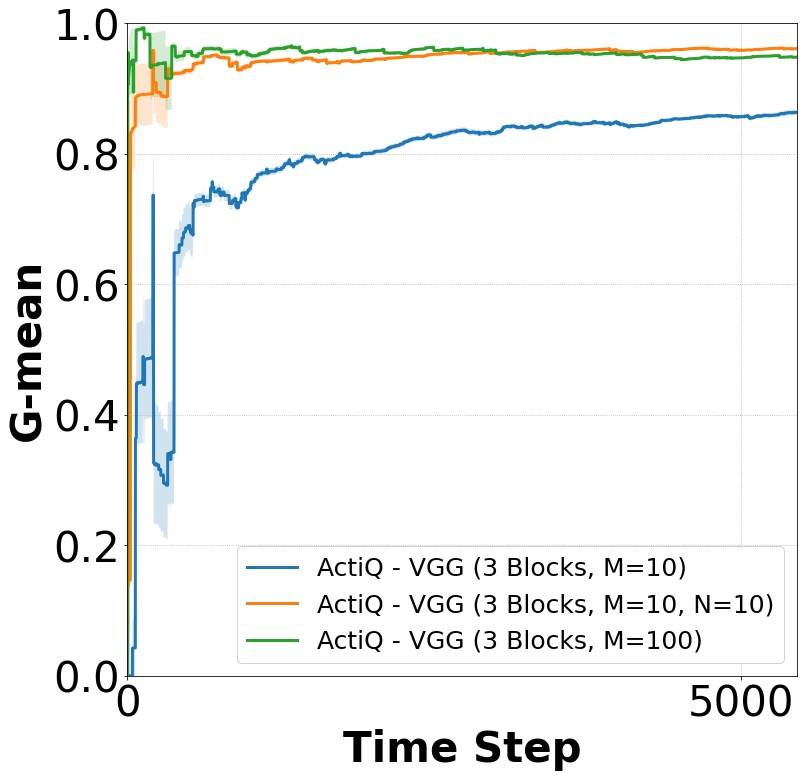}
        \caption{ActiQ-VGG, $B=10\%$}
        \label{fig:vgg_augs_comparison_b0.1}
    \end{subfigure}
    \hfill
    \begin{subfigure}[b]{0.3\columnwidth}
        \centering
        \includegraphics[width=\linewidth]{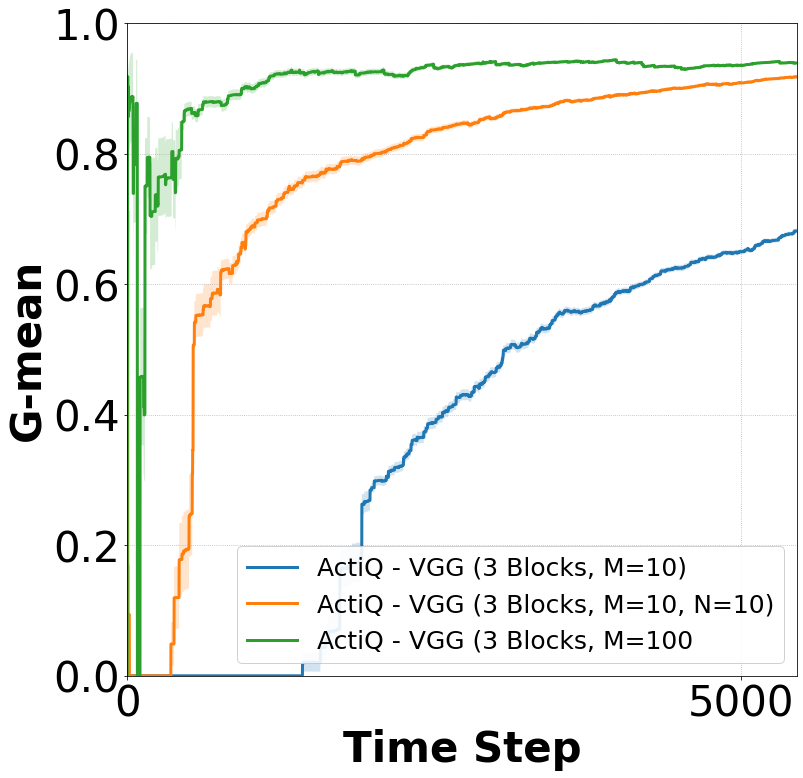}
        \caption{ActiQ-VGG, $B=1\%$}
        \label{fig:vgg_augs_comparison_b0.01}
    \end{subfigure}
    
\caption{The role of augmentation in MNIST with 1\% imbalance} % with memory $M=10$.}
\label{fig:nn_augs_comparison}
\end{figure}

\begin{figure}[t]

    \begin{subfigure}[b]{0.3\columnwidth}
        \centering
        \includegraphics[width=\linewidth]{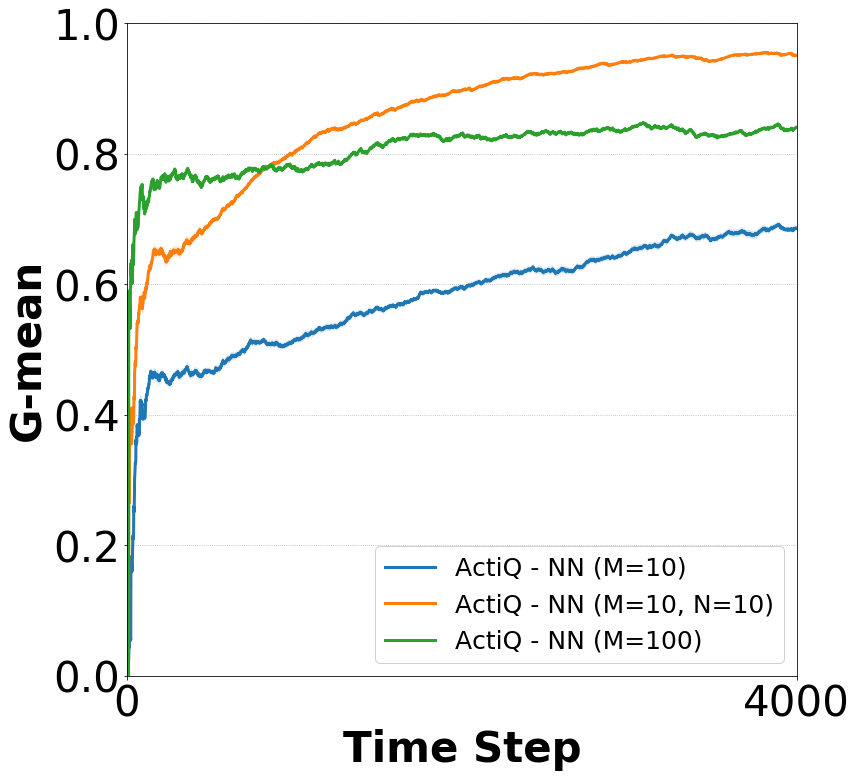}
        \caption{ActiQ-NN, $B=10\%$}
        \label{fig:nn_augs_comparison_b0.1_twoPatterns}
    \end{subfigure}
    \hfill
    \begin{subfigure}[b]{0.3\columnwidth}
        \centering
        \includegraphics[width=\linewidth]{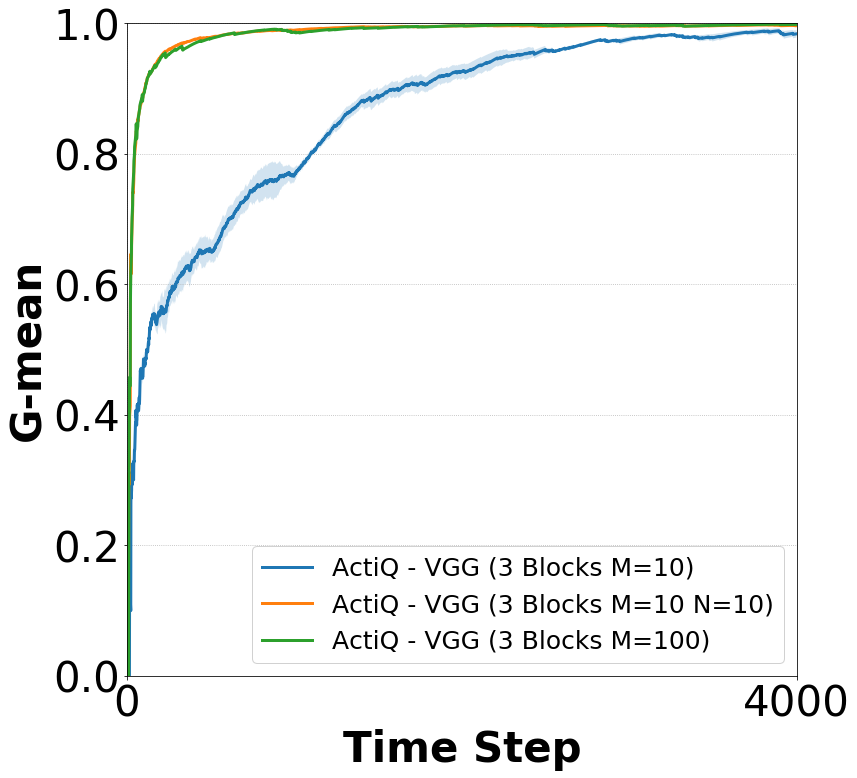}
        \caption{ActiQ-VGG, $B=10\%$}
        \label{fig:vgg_augs_comparison_b0.1_twoPatterns}
    \end{subfigure}
    \hfill
    \begin{subfigure}[b]{0.3\columnwidth}
        \centering
        \includegraphics[width=\linewidth]{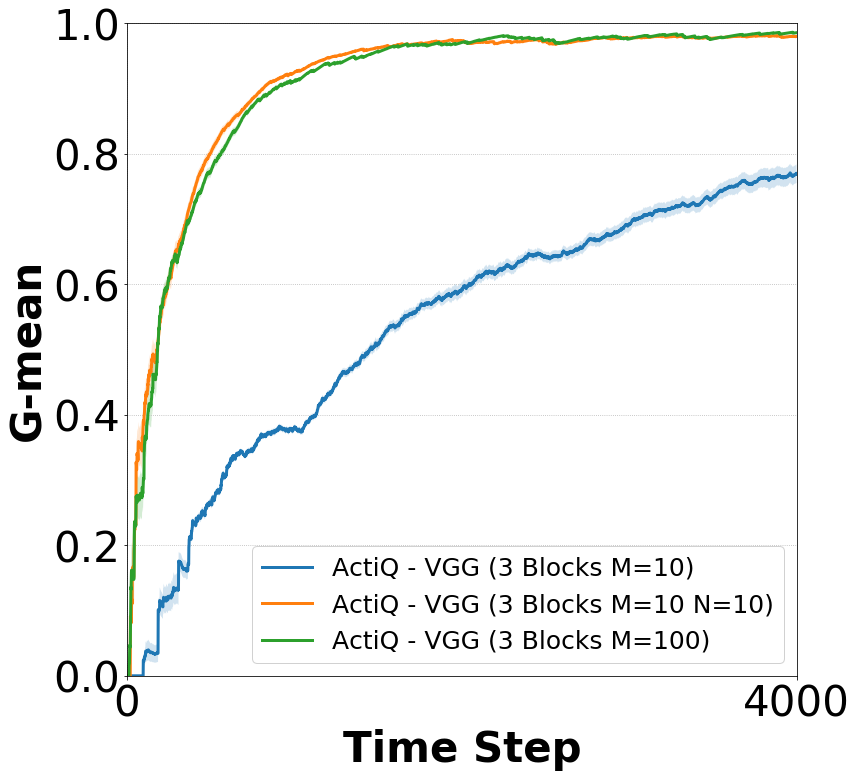}
        \caption{ActiQ-VGG, $B=1\%$}
    \label{fig:vgg_augs_comparison_b0.01_twoPatterns}
    \end{subfigure}
    
\caption{The role of augmentation in Two Patterns} % with memory $M=10$.}
\label{fig:nn_augs_comparison_twoPatterns}
\end{figure}

\begin{figure}[t]

    \begin{subfigure}[b]{0.3\columnwidth}
        \centering
        \includegraphics[width=\linewidth]{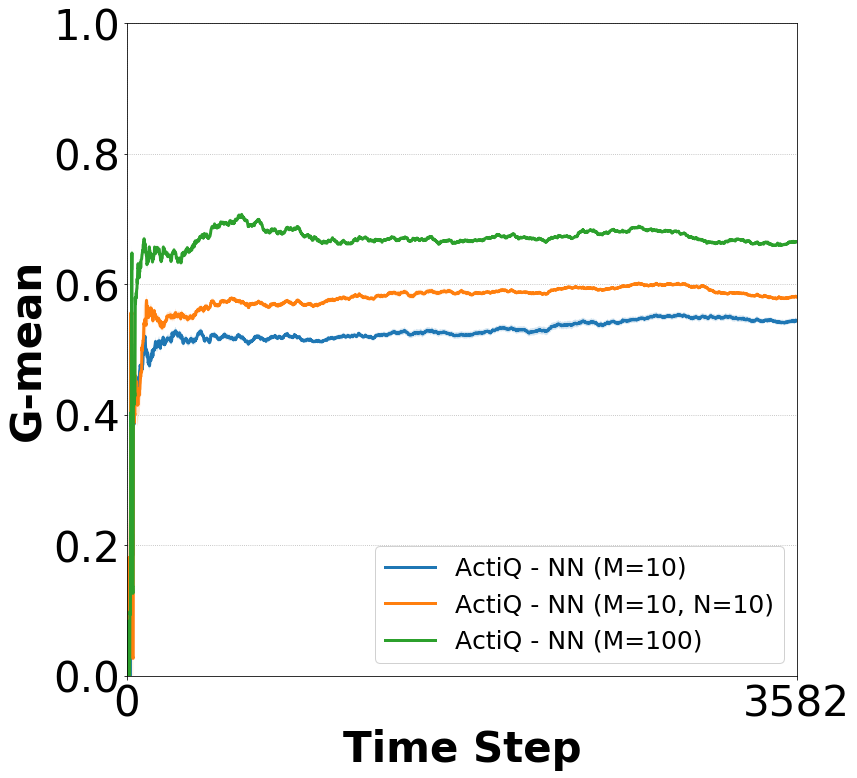}
        \caption{ActiQ-NN, $B=10\%$}
        \label{fig:nn_augs_comparison_b0.1_UWaveGestureZ}
    \end{subfigure}
    \hfill
    \begin{subfigure}[b]{0.3\columnwidth}
        \centering
        \includegraphics[width=\linewidth]{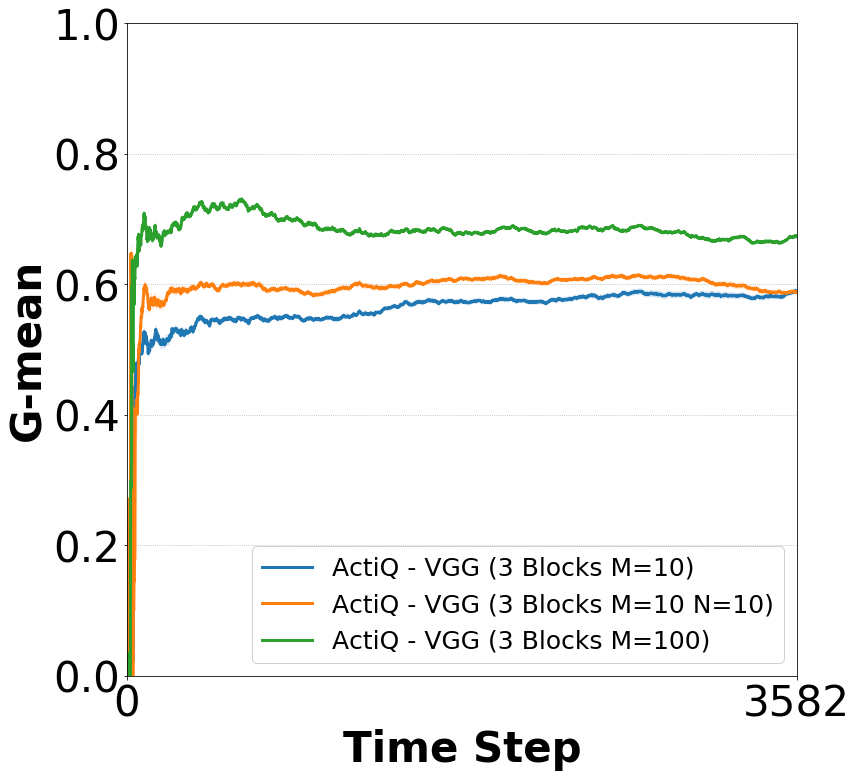}
        \caption{ActiQ-VGG, $B=10\%$}
        \label{fig:vgg_augs_comparison_b0.1_UWaveGesture}
    \end{subfigure}
    \hfill
    \begin{subfigure}[b]{0.3\columnwidth}
        \centering
        \includegraphics[width=\linewidth]{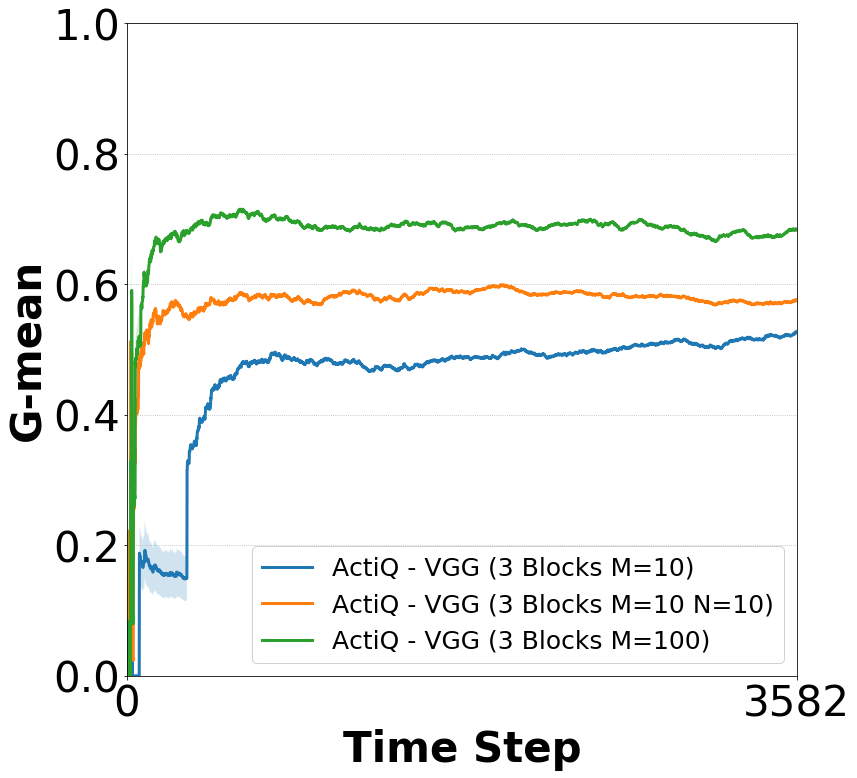}
        \caption{ActiQ-VGG, $B=1\%$}
    \label{fig:vgg_augs_comparison_b0.01_UWaveGesture}
    \end{subfigure}
    
\caption{The role of augmentation in uWave Gesture Library Z} % with memory $M=10$.}
\label{fig:nn_augs_comparison_uWave}
\end{figure}

We consider the datasets with 1\% and 10\% budget, and memory sizes of 10 and 100. The performance results of NN and the VGG (3 blocks) are presented in Figures \ref{fig:nn_augs_comparison}, \ref{fig:nn_augs_comparison_twoPatterns} and \ref{fig:nn_augs_comparison_uWave}. The augmentation techniques applied are described in Section \ref{sec:data_augmentation} and their value ranges are provided in our code. %Table \ref{tab:data_augmentation_values} and \ref{tab:data_augmentation_values_time_series}.

% \begin{figure}[t]

%     \begin{subfigure}[b]{0.49\columnwidth}
%         \centering
%         \includegraphics[width=\linewidth]{exps_plots/augmentation_comparison_nn_imbalance_1_perc_learning_curve_plot.png}
%         \caption{Imbalanced 1\%}
%         \label{fig:nn_augs_comparison_imb0.01}
%     \end{subfigure}
%     \hfill
%     \begin{subfigure}[b]{0.49\columnwidth}
%         \centering
%         \includegraphics[width=\linewidth]{exps_plots/augmentation_comparison_nn_imbalance_10_perc_learning_curve_plot.png}
%         \caption{Imbalanced 10\%}
%         \label{fig:nn_augs_comparison_imb0.1}
%     \end{subfigure}
    
% \caption{Augmentation using NN with $M=10$ and $B=10\%$}
% \label{fig:nn_augs_comparison}
% \end{figure}

% \begin{figure}[t]

%     \begin{subfigure}[b]{0.49\columnwidth}
%         \centering
%         \includegraphics[width=\linewidth]{exps_plots/augmentation_comparison_vgg_imbalance_1_perc_learning_curve_plot.png}
%         \caption{Imbalanced 1\%}
%         \label{fig:vgg_augs_comparison_imb0.01}
%     \end{subfigure}
%     \hfill
%     \begin{subfigure}[b]{0.49\columnwidth}
%         \centering
%         \includegraphics[width=\linewidth]{exps_plots/augmentation_comparison_vgg_imbalance_10_perc_learning_curve_plot.png}
%         \caption{Imbalanced 10\%}
%         \label{fig:vgg_augs_comparison_imb0.1}
%     \end{subfigure}
    
% \caption{Augmentation using VGG with $M=10$ and $B=10\%$}
% \label{fig:vgg_augs_comparison}
% \end{figure}

In Fig. \ref{fig:nn_augs_comparison_b0.1}, we compare the performance of the NN when we apply data augmentations to the MNIST samples. As a baseline, we use the NN with memory 10 without any augmentations, and we also include the NN with memory 100 with no augmentations. We notice that the use of data augmentation techniques yields slightly improved results when compared to the NN (M=10). We attribute this to the small capacity of the NN. The analogous plots for VGG (3 blocks) are shown in Figs. \ref{fig:vgg_augs_comparison_b0.1} and \ref{fig:vgg_augs_comparison_b0.01}. It is evident that data augmentations are very effective when applied to the VGG network. For both MNIST variations, the performance of the model with augmented queues of size 10, reaches the performance of the models with memory size 100, which is a significant improvement.

In Fig. \ref{fig:nn_augs_comparison_b0.1_twoPatterns}, we examine NN's performance when we apply augmentations to the Two Patterns samples. Again, as a baseline we consider the NN with memory 10 and 100, without any augmentation. It is evident, that data augmentation helped memory size 10 to outperform both memory size 10 and 100, without augmentation. Similarly, for VGG in Figs. \ref{fig:vgg_augs_comparison_b0.1_twoPatterns} and \ref{fig:vgg_augs_comparison_b0.01_twoPatterns} we notice that augmentation increases the VGG's performance compared to memory size 10 without augmentation. Importantly, in both VGG experiments, augmentation achieves similar performance as the memory size 100.

Fig. \ref{fig:nn_augs_comparison_uWave} shows the NN and VGG performance results on the uWave Gesture Library Z. We observe that for budget 10\% as shown in Figs \ref{fig:nn_augs_comparison_b0.1_UWaveGestureZ} and \ref{fig:vgg_augs_comparison_b0.1_UWaveGesture}, augmentation can slightly improve over memory 10 without augmentation. In Fig \ref{fig:vgg_augs_comparison_b0.01_UWaveGesture} with budget 1\%, Augmented Queues performance increased significantly when compared with memory 10 without augmentations.

Important remarks are as follows:
\begin{itemize}
    \item Augmentation has a drastic improvement on model performance. This is attributed to the increase of the training set, which appears to have a similar effect as if the original memory was increased. For the shallow model NN, an improvement was observed but to a lesser degree.
    
    \item The proposed method uses less space, as augmentation happens on-the-fly at training times only; augmented data do not reserve memory when they are not in use.
\end{itemize}

\section{Conclusion and Future Work}\label{sec:conclusion}
Learning online poses major challenges which hinder the deployment of learning models. We introduced Augmented Queues which addresses the problems of limited labelled data and class imbalance. Augmented Queues synergistically combines on-the-fly data augmentation with active learning. We demonstrate its applicability using image and time-series augmentations, and we show that it significantly improves the learning quality and speed. Future work will examine:

\textbf{Non-stationary environments}. While this work has focused on the critical challenges of limited label availability, class imbalance, and limited storage, other challenges exist, such as, concept drift \cite{ditzler2015learning, gama2014survey}. %Despite that MNIST exhibits nonstationarity as new different handwriting styles are observed, future work will thoroughly examine the Augmented Queues method under various characteristics of drift.

% \textbf{Deeper neural models}. The scalability of Augmented Queues will be examined when applying it in conjunction with deeper models, e.g., ResNet \cite{he2016deep} and Transformers \cite{vaswani2017attention}.

\textbf{Augmentation in the latent space}. One promising direction is to perform augmentation in the latent space, e.g., using Siamese networks \cite{malialis2022nonstationary}. This would be challenging as the latent space changes over time due to online incremental learning.

\Urlmuskip=0mu plus 1mu\relax
\bibliographystyle{IEEEtran}
\bibliography{wcci_ijcnn_22}

\end{document}